\documentclass[conference]{IEEEtran}
\IEEEoverridecommandlockouts
\usepackage{cite}
\usepackage{amsmath,amssymb,amsfonts}
\usepackage{algorithmic}
\usepackage{graphicx}
\usepackage{textcomp}
\usepackage{xcolor}
\renewcommand{\figurename}{Figure}
\renewcommand{\thetable}{\arabic{table}}
\usepackage{caption}
\usepackage{fancyhdr}
\def\BibTeX{{\rm B\kern-.05em{\sc i\kern-.025em b}\kern-.08em
    T\kern-.1667em\lower.7ex\hbox{E}\kern-.125emX}}

\fancypagestyle{firstpage}{%
 \fancyhf{}%
 \fancyhead[L]{2024 2nd International Conference on Information and Communication Technology (ICICT)\newline
  \textit{October 21-22, Dhaka, Bangladesh}}
  
 \fancyfoot[L]{%
  \normalsize{979-8-3315-0822-7/24/\$31.00 ~\copyright2024 IEEE}
 }%
}
\begin{document}

\title{English offensive text detection using CNN based Bi-GRU model\\

}

\author{\IEEEauthorblockN{
 Tonmoy Roy\textsuperscript{1}
}
\IEEEauthorblockA{\textit{Data Analytics \& Information Systems 
} \\
\textit{Utah State University}\\
Utah, United States \\
tonmoy.roy@usu.edu
}
\and
\IEEEauthorblockN{Md Robiul Islam\textsuperscript{2}}
\IEEEauthorblockA{\textit{Computer Science} \\
\textit{William \& Mary}\\
Virginia, United States \\
robiul.cse.uu@gmail.com}
\and
\IEEEauthorblockN{Asif Ahammad  Miazee\textsuperscript{3}}
\IEEEauthorblockA{\textit{Computer Science} \\
\textit{Maharishi International University}\\
Iowa, United States \\
asifahammad7@gmail.com}
\and
\IEEEauthorblockN{Anika Antara\textsuperscript{4}}
\IEEEauthorblockA{\textit{Electrical and Electronics Engineering} \\
\textit{Brac University}\\
Dhaka, Bangladesh \\
anikaantara1000@gmail.com}
\and
\IEEEauthorblockN{Al Amin\textsuperscript{5}}
\IEEEauthorblockA{\textit{Computer Science } \\
\textit{Uttara University}\\
Dhaka, Bangladesh \\
alaminbhuyan321@gmail.com}
\and
\IEEEauthorblockN{Sunjim Hossain\textsuperscript{6}}
\IEEEauthorblockA{\textit{Computer Science} \\
\textit{Northern University}\\
Dhaka, Bangladesh \\
hossainsunjim@gmail.com
}
}

\maketitle
\thispagestyle{firstpage}

\begin{abstract}
Over the years, the number of users of social media has increased drastically. People frequently share their thoughts through social platforms, and this leads to an increase in hate content. In this virtual community, individuals share their views, express their feelings, and post photos, videos, blogs, and more. Social networking sites like Facebook and Twitter provide platforms to share vast amounts of content with a single click. However, these platforms do not impose restrictions on the uploaded content, which may include abusive language and explicit images unsuitable for social media. To resolve this issue, a new idea must be implemented to divide the inappropriate content. Numerous studies have been done to automate the process. In this paper, we propose a new Bi-GRU-CNN model to classify whether the text is offensive or not. The combination of the Bi-GRU and CNN models outperforms the existing models.
\end{abstract}


\section{Introduction}
Hate speech is a distinct form of language that involves abusive behaviour. The targets of this phenomenon are specifically selected based on their personal qualities or demographic background, including race, ethnicity, religion, colour, sexual orientation, or other comparable criteria \cite{nobata2016abusive}. A multitude of academics are diligently focused on addressing the issue of hate speech detection using natural language processing techniques. They are developing practical frameworks and creating automatic classifiers that rely on supervised machine learning models \cite{fortuna2018survey}. 
Sharing inappropriate content on social media platforms like Twitter and Facebook has become increasingly effortless, often targeting specific individuals or groups. Additionally, toxic language can manifest in various forms, including cyberbullying, which has played a key role in contributing to suicide. \cite{hinduja2010bullying}. According to the United Nations strategy and plan of action on hate speech, there is no internationally agreed-upon legal definition, However, it involves incitement, which means intentionally encouraging discrimination, hostility, and violence\cite{Un}. 

Similarly, offensive speech refers to writing that includes abusive slurs or degrading terms \cite{gaydhani2018detecting}, which are often mistaken for hate speech in various situations. Offensive language and hate speech detection are two specialised areas of study within the subject of natural language processing. The primary obstacle is in the fact that the majority of unsuitable content found online is presented in the form of natural language text. Consequently, it is imperative to develop efficient tools capable of extracting and analysing this content from unorganised textual data.

These technologies often utilise methods that are based on natural language processing (NLP), retrieval of data, machine learning, and deep learning. Research conducted by \cite{hu2015protecting} focuses on safeguarding children from inappropriate content in mobile applications. The study suggests techniques for parents to assess the maturity level of smartphone apps, allowing them to select apps that are appropriate for their children's ages. Unfortunately, it is not practical to manually filter out harmful content on a large scale, and manually identifying or removing anything from the internet is a laborious undertaking. This serves as a driving force for researchers to develop automated techniques that can identify offensive information on social media platforms.

This study presents comprehensive experiments aimed at resolving the problem of categorising improper content through the utilisation of machine learning and neural network models. The methodology we employ involves optimising hyperparameters and utilising word embedding features on a dataset in the English language. Our main contribution in this paper as follows:
\begin{itemize}
     \item We proposed a fine-tuned 1D Convolutional Neural Network (CNN) with Bi-GRU based offensive text detection model, which outperforms other benchmark models.
     \item The dataset contains more than 31 thousands tweets which is a big data to do work.
\end{itemize}

The paper is organized as follows: In section II, we review the relevant literature on classifying offensive speech. Section III details our proposed methodology. Section IV covers our experiments and results. In Section V, we discuss the challenges encountered during implementation. Finally, in section VI, we conclude with a summary and outline future work.

\section{Related Works}
Social life has become part of life nowadays. According to the source, on average, a user spends 3 hours and 15 minutes on their phone each day, and that number increases if there is a holiday \cite{timespent}. Individuals check their phones 58 times within 24 hours. Over the past few years, there has been a rapid development of social media, resulting in both benefits and drawbacks in its utilisation. Hate speech commonly targets specific races, religions, sexual orientations, genders, and castes, among other categories. This problem has prompted numerous academics to investigate an approach capable of identifying hate speech that is targeted towards specific individuals or groups \cite{ibrohim2023hate}.

Machine learning techniques for natural language processing (NLP) have traditionally relied on shallow models like Support Vector Machines (SVM) and Logistic Regression (LR). These models are trained using features that are both high-dimensional and sparse. Most techniques primarily concentrate on extracting textual aspects. Some scholars have employed lexical features, such as dictionaries \cite{alshari2018effective} and bag-of-words \cite{pandey2022hate}. Kumari et. al. propose a hybrid model for identifying aggressive posts that include both images and text on social media platforms \cite{kumari2021multi}. In contrast, Kovacs et. al. employ various machine learning techniques and deep learning models to automatically detect speech that is both hateful and offensive \cite{kovacs2021challenges}. Although machine learning (ML) has encountered problems, there have been significant advancements in developing ML-based systems for automatic detection, which have yielded promising outcomes. Some noteworthy examples of classifiers are the Naive Bayes (NB) \cite{pandey2022hate}, Logistic Regression (LR) \cite{oriola}, and Support Vector Machines (SVM) \cite{oriola}. Offensive speech detection is not only used in English; it is also used in Hindi as well \cite{vashistha2020online}. Also in Spanish and Italian \cite{jiang2021cross}. Many researchers work on several languages to detect offensive text \cite{deshpande2022highly}. Corazza et al. \cite{corazza2020multilingual} employ datasets in three distinct languages (English, Italian, and German) and train various models including LSTMs, GRUs, Bidirectional LSTMs, and others. Huang et al. \cite{huang2020multilingual} created a multilingual Twitter hate speech dataset by combining data from 5 languages. They also added demographic information to investigate the presence of demographic bias in hate speech classification. Aluru et al. \cite{aluru2020deep} employed datasets from 8 different languages and produced embeddings using LASER \cite{laser} and MUSE \cite{muse}. These embeddings were inputted into various architectures, such as CNNs, GRUs, and different Transformer models. While the study achieved satisfactory performance across these languages, it had two main shortcomings. Firstly, the claim that the models are generalizable is questionable, as they did not leverage datasets from multiple other languages and lacked fine-tuning of parameters for better generalization. Consequently, these models are unlikely to perform well outside of the 8 languages studied. Secondly, the research focused primarily on models suited for low-resource settings without addressing their performance in resource-abundant environments. By aggregating datasets from 11 different languages, his study achieved superior performance \cite{islam2024enhancing}.

\section{Methodology}
In this section, we present our proposed methodology for offensive text classification. Our proposed Bi-GRU and CNN models take input text and output the probability of it being in an inappropriate class. Our model consists of nine layers (a) Input Layer, (b) Embedding Layer, (c) Convolutional Layer(Conv1D), (d) MaxPooling Layer, (e) Bi-directional GRU\_1(Bi-GRU), (f) Bi-directional GRU\_2(Bi-GRU), (g) Dense Layer, (h) Dropout Layer, (i) Dense Layer.
\begin{figure*}[]
    \centering
    \includegraphics[width=.8\textwidth, height=0.5\textwidth]{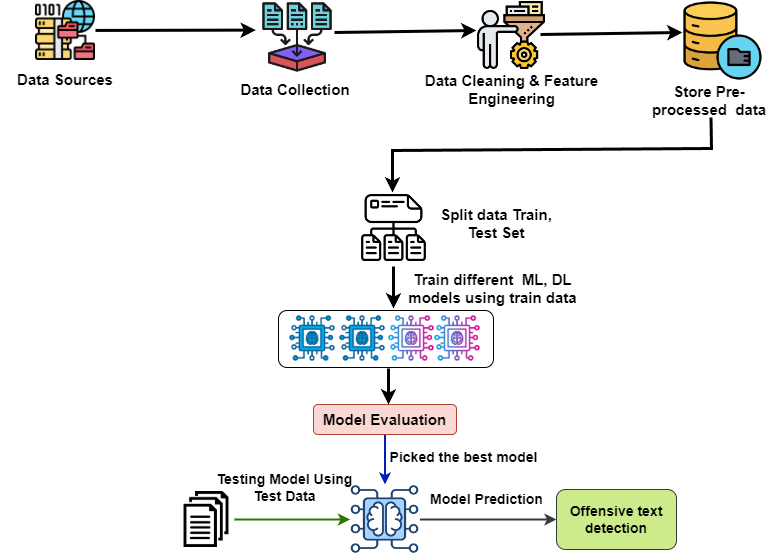}
    \caption{Proposed methodology}
    \label{fig:methodology}
\end{figure*}

\subsection{Dataset Descriptions}
In this subsection, we present our dataset that we have used in our experiments. The dataset contains 31,962 tweets. This dataset has been split into a training set and a test set. The ratio is 80:20 for training and testing respectively.

    \begin{table}[ht]
    \caption{Dataset}\label{table:dataset}
    \centering
    \begin{tabular}{|p{4cm}|p{4cm}|}
    \hline
    \textbf{Classes} & \textbf{Total Tweet}  \\ [1ex]
    \hline
    Offensive & 29720 \\ [1ex]
    \hline
   Not offensive & 2242 \\ [1ex]
    \hline
    \end{tabular}
    \end{table}

\subsection{Pre-processing}
In order to enhance the efficiency of our approach, it is important to carry out pre-processing steps to cleanse our textual data. Initially, the tweet underwent a process where all numerical values, punctuation marks, URLs (starting with http:// or www.), and symbols (like emojis, hashtags, and mentions) were eliminated. This was done because these elements do not contribute to the sentiment-related content of the tweet. The tweets were first broken down into individual words or phrases, a process called tokenization. This was done using a tool from the NLTK library. Then, all common words with little meaning on their own (like "the", "a", "is") were removed from the tokens. These common words are also provided by the NLTK library.

\subsection{Bi-directional GRU (Bi-GRU)}
Regular GRU models only consider information from previous words in a sequence. But to truly grasp the meaning of a word, it's important to also understand what comes after it. That's where Bidirectional GRUs (BiGRUs) come in. Our system uses BiGRUs, which are essentially two GRUs working together. One reads the text forward, the other backward. This allows the BiGRU to capture important details from the text and analyze each word in the context of both its past and future neighbors. This gives BiGRUs an edge over traditional machine learning models when it comes to understanding the nuances of language and detecting hate speech.

\begin{figure}[ht]
    \centering
    \includegraphics[scale=0.5]{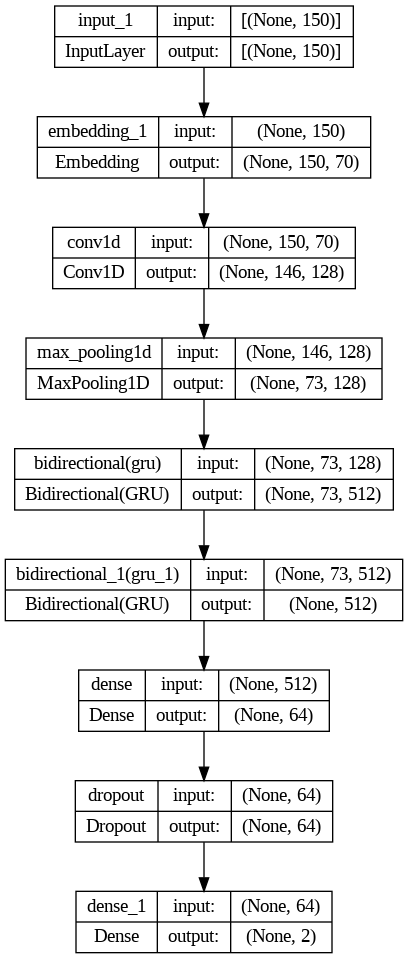}
    \caption{Model Flow Chart}
    \label{fig:model_flowchart}
\end{figure}

\subsection{Experimental setup}
For these studies, we utilise Keras and Tensorflow as the underlying frameworks. We utilise Google Colab to execute the tests with the assistance of the Graphical Processing Unit (GPU). Regarding the training process, we employ a total of 100 epochs. Several traditional machine learning techniques were employed, including Naive Bayes, Support Vector Machine, Logistic Regression, Random Forest, and Adaboost. Furthermore, alongside these models, various techniques for word representation were employed, including count vectors as features and term frequency inverse document frequency (TF-IDF).






\section{Experimentations and results}
In this section, we discuss the model's performance. We discuss about accuracy, precision and recall.

\subsection{Results}
Every model undergoes training using the training dataset, and its performance is assessed using standard classification metrics: Accuracy (Acc), F1-score (F1), Recall (R), and Precision (P). In order to improve the outcomes, different hyperparameters were experimented with and integrated \cite{islam2024phishguard}. This paper provides a comprehensive overview of all the findings, with a particular emphasis on the models that performed the best. Multiple preliminary evaluations were conducted prior to submitting the final results.

\begin{figure}[ht]
    \centering
    \includegraphics[width=0.5\textwidth, height=0.37\textwidth]{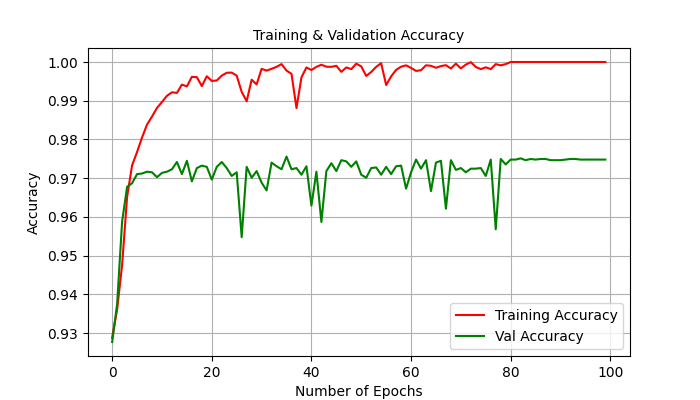}
    \caption{Training and Validation Accuracy}
    \label{fig:accuracy}
\end{figure}

\begin{figure}[ht]
    \centering
    \includegraphics[width=0.5\textwidth, height=0.37\textwidth]{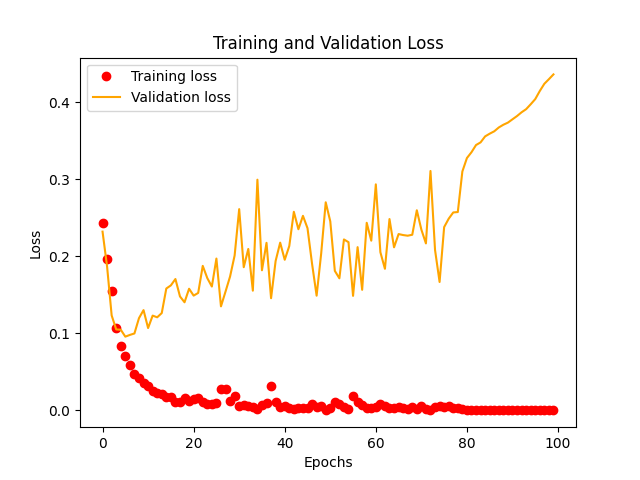}
    \caption{Training and Validation Loss per epoch}
    \label{fig:loss}
\end{figure}

\begin{figure}[ht]
    \centering
    \includegraphics[width=0.5\textwidth, height=0.37\textwidth]{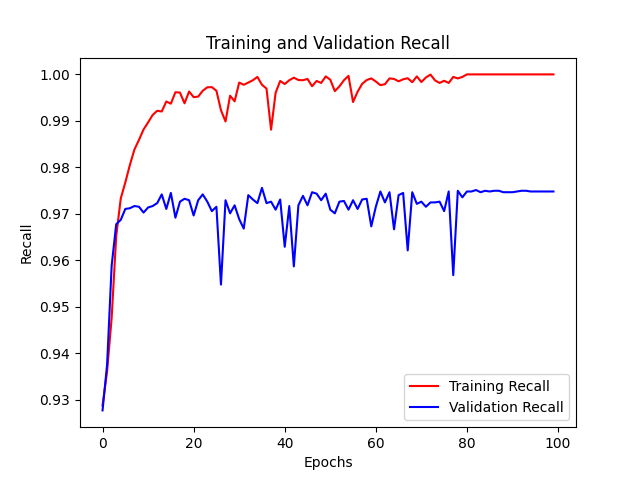}
    \caption{Training and Validation Recall}
    \label{fig:recall}
\end{figure}

\begin{figure}[ht]
    \centering
    \includegraphics[width=0.5\textwidth, height=0.37\textwidth]{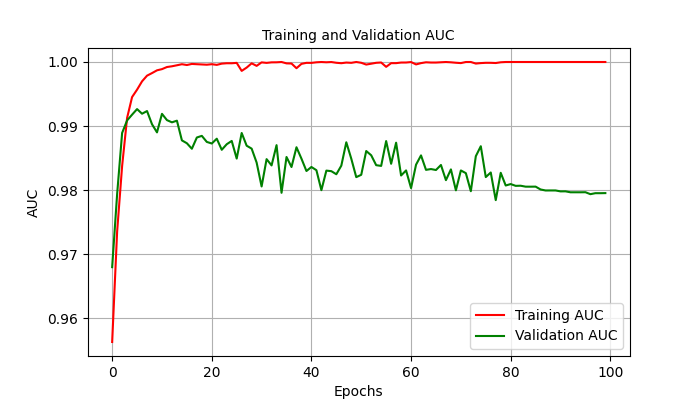}
    \caption{Training and Validation AUC}
    \label{fig:auc}
\end{figure}

Fig. \ref{fig:accuracy} shows the training and validation accuracy of the proposed model. Fig. \ref{fig:loss} depicts the suggested model loss graph for 100 epochs on training and validation stage. Beside that Fig. \ref{fig:recall} and Fig. \ref{fig:auc} represents the training and validation recall and AUC graph respectively. Table \ref{table:MLModelAcc} shows the evaluation results of different machine learning models and Table \ref{table:modelCompare} shows the comparative analysis of some other existing research.


    \begin{table}[ht]
    \caption{Model Evaluation table}\label{table:MLModelAcc}
    \centering
    \begin{tabular}{|c|p{2cm}|c|c|c|c|}
    \hline
    \textbf{No.} & \textbf{Algorithm} & \textbf{Accuracy} & \textbf{Precision} & \textbf{Rcall} & \textbf{F1-Score} \\ [1.5ex]
    \hline
    1 & Logistic Regression & 96.86 & 86.50 & 51.27& 64.21 \\ [1.5ex]
    \hline
    2 & RandomForest Classifier & 96.56 & 84.33 & 54.26 & 67.56 \\ [1.5ex]
    \hline
    3 & MultinomialNB & 95.97 & 95.65 & 95.97 & 95.47\\ [1.5ex]
    \hline
    4 & KNeighbors Classifier & 95.97 & 81.41 & 81.41 & 81.41 \\ [1.5ex]
    \hline
    5 & Linear SupportVectorClassification & 94.90 & 98.46 & 82.05 & 89.51 \\ [1.5ex]
    \hline
    \end{tabular}
    \end{table}
    
    \begin{table}[ht]
    \caption{Research comparison with others}\label{table:modelCompare}
    \centering
    \begin{tabular}{|c|p{1cm}|c|c|c|}
    \hline
    \textbf{No.} & \textbf{Authors}  &   \textbf{Algorithms} & \textbf{FE method} & \textbf{Accuracy} \\ [1ex]
    \hline
    1 & Vashistha et. al. \cite{vashistha2020online} & LR & Word embedding & 76.90\% \\ [1ex]
    \hline
    2 & Aluru et. al. \cite{aluru2020deep} & LR & MUSE and LASER & 76.90\% \\ [1ex]
    \hline
     3 & Deshpande et. al. \cite{deshpande2022highly} & TL & mBERT & 76.90\% \\ [1ex]
    \hline
     4 & Proposed model & Bi-GRU and CNN & CountVectorizer & 96.97\% \\ [1ex]
    \hline
    
    \end{tabular}
    \end{table}
\section{Conclusion and Future work}
This research describes a system that combines Convolutional Neural Networks (CNN) and Bidirectional Gated Recurrent Units (Bi-GRU). We were motivated to create contextual embeddings by the use of social networks, namely by utilising a Twitter dataset. We use the acquired information from this language model designed to detect offensive language and expressions of hate in written content. We conducted an assessment of various supervised machine learning classifiers to detect nasty and abusive content on Twitter, using a dataset specifically collected from Twitter. Deep learning enables the automatic acquisition of multi-level feature representations, while typical machine learning-based NLP systems rely extensively on manually engineered features. Creating these artisanal elements requires a significant amount of time and effort and may not always be fully developed. The deep learning models achieved outstanding results in reducing false positives and showed promising outcomes in decreasing false negatives during the classification process. Our objective is to do more investigation into the application of deep neural network architectures for the purpose of detecting hate speech. We will conduct our inquiry by employing a range of word embedding approaches. In addition, we intend to broaden our research by incorporating additional datasets, particularly those in the Bangla language \cite{amin2024sentiment}.

\bibliographystyle{ieeetr}
\bibliography{references}

\end{document}